\documentclass[10pt,twocolumn,letterpaper]{article}
\pdfoutput=1
\usepackage{cvpr}
\usepackage{times}
\usepackage{epsfig}
\usepackage{graphicx}
\usepackage{amsmath}
\usepackage{amssymb}
\usepackage{url}
\usepackage{multirow}
\usepackage{arydshln}

\newcommand{\chapternote}[1]{{%
  \renewcommand{\thefootnote}{\fnsymbol{footnote}}%
  \footnotetext[0]{#1}
}}

\usepackage[breaklinks=true,bookmarks=false]{hyperref}

\cvprfinalcopy 

\def\cvprPaperID{2715} 

\ifcvprfinal\fi
\begin{document}

\title{Deep Affordance-grounded Sensorimotor Object Recognition}
\author{Spyridon Thermos$^{1,2}$ \quad Georgios Th. Papadopoulos$^{1}$ \quad Petros Daras$^{1}$ \quad Gerasimos Potamianos$^{2}$\\
\\
$^{1}$Information Technologies Institute, Centre for Research and Technology Hellas, Greece \\ $^{2}$Department of Electrical and Computer Engineering, University of Thessaly, Greece \\
{\tt\small \{spthermo,papad,daras\}@iti.gr} \qquad {\tt\small gpotam@ieee.org}
}

\maketitle
\thispagestyle{empty}

\begin{abstract}
It is well-established by cognitive neuroscience  that human perception of objects constitutes a complex process, where object appearance information is combined with evidence about the so-called object ``affordances", namely the types of actions that humans typically perform when interacting with them. This fact has recently motivated the ``sensorimotor" approach to the challenging task of automatic object recognition, where both information sources are fused to improve robustness. In this work, the aforementioned paradigm is adopted, surpassing current limitations of sensorimotor object recognition research. Specifically, the deep learning paradigm is introduced to the problem for the first time, developing a number of novel neuro-biologically and neuro-physiologically inspired architectures that utilize state-of-the-art neural networks for fusing the available information sources in multiple ways. The proposed methods are evaluated using a large RGB-D corpus, which is specifically collected for the task of sensorimotor object recognition and is made publicly available. Experimental results demonstrate the utility of affordance information to object recognition, achieving an up to 29\% relative error reduction by its inclusion.
\end{abstract}
\chapternote{The work presented in this paper was supported by the European Commission under contract H2020-687772 MATHISIS}
\section{Introduction} \label{intro}
Object recognition constitutes an open research challenge of broad interest in the field of computer vision. Due to its impact in application fields such as office automation, identification systems, security, robotics, and the industry, several research groups have devoted intense efforts in it (see for example the review in \cite{survey}). However, despite the significant advances achieved in the last decades, satisfactory performance in real-world scenarios remains a challenge. One plausible reason is the sole use of static object appearance features \cite{donahue, liang, liu}. Such cannot sufficiently handle the object appearance variance, occlusions, deformations, and illumination variation.

Research findings in cognitive neuroscience establish that object recognition by humans exploits previous experiences of active interaction with the objects of interest. In particular, object perception is based on the fusion of sensory (object appearance) and motor (human-object interaction) information. Central role in this so-called ``sensorimotor object recognition" theory has the notion of object affordances. According to Gibson \cite{gibson1}, ``the affordances of the environment are what it offers the animal", implying the complementarity between the animal and the environment. Based on this theory, Minsky \cite{minsky} argues on the significance of classifying items according to what they can be used for, \ie what they afford. These theoretical foundations have resulted to the so-called function-based reasoning in object recognition, which can be viewed as an approach applicable to environments in which objects are designed or used for specific purposes \cite{rivlin}. Moreover, the work in \cite{suttonAff} describes three possible ways for extracting functional (affordance) information for an object: a) ``Function from shape", where the object shape provides some indication of its function; b) ``Function from motion", where an observer attempts to understand the object function by perceiving a task being performed with it; and c) ``Function from manipulation", where function information is extracted by manipulating the object. The present work focuses on (b). 

Concerning the neuro-physiological and the corresponding cognitive procedures that take place in the human brain during sensorimotor object recognition, it is well established that there are two main streams that process the visual information \cite{neuro1}: a) the dorsal, which projects to the posterior parietal cortex and is involved in the control of actions (motor), and b) the ventral, which runs to the inferotemporal cortex and is involved in the identification  of the objects (sensory). There is accumulated evidence that the two streams interact at different information processing stages \cite{neuro2}: a) computations along the two pathways proceed both independently and in parallel, reintegrating within shared target brain regions; b) processing along the separate pathways is modulated by the existence of recurrent feedback loops; and c) information is transferred directly between the two pathways at multiple stages and locations along their trajectories. These identified interconnections indicate how the human brain fuses sensory and motor information to achieve robust recognition. Motivated by the above, mimicking the human sensorimotor information processing module in object recognition by machines may hold the key to address the weaknesses of current systems.

Not surprisingly, affordance-based information has already been introduced into the object recognition problem. However, current systems have been designed based on rather simple classification, fusion, and experimental frameworks, failing to fully exploit the potential of the affordance stream. In particular, these works have not yet exploited the recent trend in computer vision of employing very deep Neural Network (NN) architectures, the so-called ``Deep Learning" (DL) paradigm. DL methods outperform all previous hand-crafted approaches by a large margin \cite{alex,google,zeiler}.

In this paper, the problem of sensorimotor 3D object recognition is investigated using DL techniques. The main contributions lie in the:

\begin{itemize}
\item \textbf{Design of novel neuro-biologically and neuro-physiologically grounded neural network architectures} for sensorimotor object recognition, exploiting the state-of-the-art automatic feature learning capabilities of DL techniques; to the best of the authors' knowledge, this is the first work that introduces the DL paradigm to sensorimotor object recognition.

\item \textbf{NN-based implementation of multiple recent neuro-scientific findings} for fusing the sensory (object appearance) and motor (object affordances) information streams in a unified machine perception computational model. Until now, such neuro-scientific findings have not been transferred to computer vision systems.

\item \textbf{Large number of complex affordance types} supported by the proposed methodology. In particular: a) a significantly increased number of affordances compared to current works that only use few (up to 5); b) complex types of object affordances (\eg squeezable, pourable) that may lead to complex object manipulations or even significant object deformations, compared to the relatively simple binary ones currently present in the literature (\eg graspable, pushable); and c) continuous-nature affordance types, moving beyond plain binary analysis of presence/non-presence of a given affordance, while modeling the exact dynamics of the exhibited affordances.

\item Introduction of a \textbf{large, public RGB-D object recognition dataset}, containing several types of interactions of human subjects with a set of supported objects. This is the first publicly available sensorimotor object recognition corpus, including 14 classes of objects, 13 categories of affordances, 105 subjects, and a total number of approximately 20,8k human-object interactions. This dataset yields sufficient data for training DL algorithms, and it is expected to serve as a benchmark for sensorimotor object recognition research.

\item \textbf{Extensive quantitative evaluation} of the proposed fusion methods and comparison with traditional probabilistic fusion approaches. 

\end{itemize}

The remainder of the paper is organized as follows: Section \ref{related} presents related work in the field of sensorimotor object recognition. Section \ref{data} discusses the introduced 3D object recognition dataset. Section \ref{sensorimotor} details the designed NN architectures. Section \ref{results} presents the experimental results, and Section \ref{conclusion} concludes this paper.

\section{Related Work} \label{related}

Most sensorimotor object recognition works have so far relied on simple fusion schemes (e.g using simple Bayesian models or the product rule), hard assumptions (\eg naive Gaussian prior distributions), and simplified experimental settings (\eg few object types and simple affordances). In particular, Kjellstr\"om \etal~\cite{kragic2} model the spatio-temporal information by training factorial conditional random fields with 3 consecutive object frames and extract action features using binary SVMs; a small dataset with 6 object types and 3 affordances is employed. H\"ogman \etal~\cite{kragic1} propose a framework for interactive classification and functional categorization of 4 object types, defining a Gaussian Process to model object-related Sensori-Motor Contigencies \cite{smc} and then integrating the ``push" action to categorize new objects (using a naive Bayes classifier). Additionally, Kluth \etal~\cite{kluth} extract object GIST-features \cite{gist} and model the possible actions using a probabilistic reasoning scheme that consists of a Bayesian inference approach and an information gain strategy module. A visuo-motor classifier is implemented in \cite{castellini} in order to learn 5 different types of grasping gestures on 7 object types, by training an SVM model with object feature clusters (using K-means clustering) and a second SVM with 22 motor features (provided by a CyberGlove); the predictions are fused with a weighted linear combination of Mercer kernels. Moreover, in the field of robotics, affordance-related object recognition has relied on predicting opportunities for interaction with an object by using visual clues \cite{aldoma, hermans} or observing the effects of exploratory actions \cite{lyubova, moldovan}. 


Clearly, design and evaluation of complex data-driven machine perception systems for sensorimotor object recognition based on the state-of-the-art DL framework has not been considered in the literature. Such systems should not depend on over-simplified or hard assumptions and would target the automatic learning of the highly complex sensorimotor object recognition process in realistic scenarios.

\section{RGB-D Sensorimotor Dataset} \label{data}
In order to boost research in the field of sensorimotor object recognition, a large-scale dataset of multiple object types and complex affordances has been collected and is publicly available at {\footnotesize \url{http://sor3d.vcl.iti.gr/}}. The corpus constitutes the broadest and most challenging one in the sensorimotor object recognition literature, as summarized in Table \ref{tbl:datasets}, and can serve as a challenging benchmark, facilitating the development and efficient evaluation of sensorimotor object recognition approaches.

\begin{figure}[t]
\begin{center}
	\includegraphics[width=\linewidth]{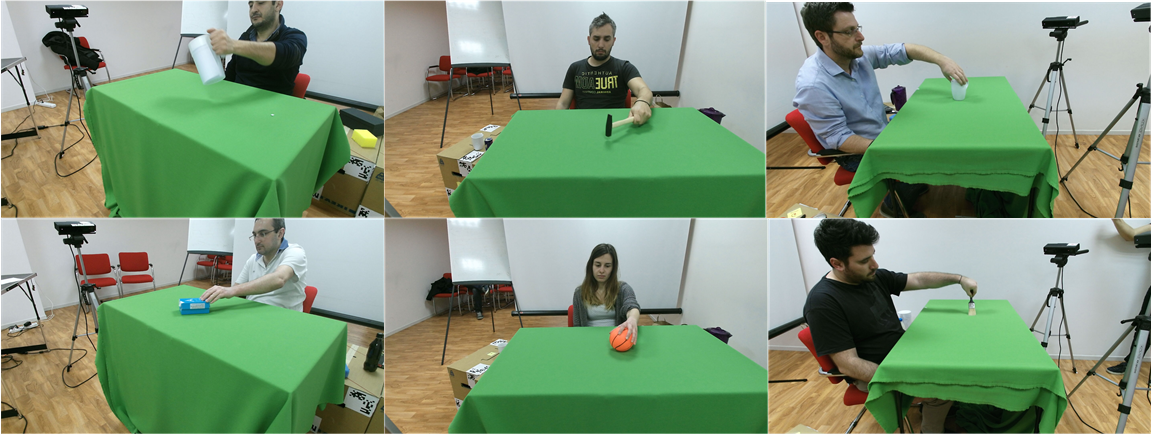}
\end{center}
   \caption{Examples of human-object interactions captured by the 3 Kinect sensors employed in the corpus recording setup.}
\label{fig:dataset}
\end{figure}

The corpus recording setup involved three synchronized Microsoft Kinect II sensors \cite{kinect} in order to acquire RGB ($1920\times 1080$ resolution) and depth ($512\times 424$ resolution) streams from three different viewpoints, all at 30 Hz frame rate and an approximate 1.5 meters ``head-to-device" distance. A monitor was utilized for displaying the ``prototype" instance before the execution of every human-object interaction. Additionally, all involved subjects were provided with a ring-shaped remote mouse, held by the other hand than that interacting with the objects. This allowed the participants to indicate by themselves the start and end of each session (\ie performing real-time annotation). Before the execution of any interaction, all objects were placed at a specific position on a desk, indicated by a marker on the table cloth. The dataset was recorded under controlled environmental conditions, \ie with negligible illumination variations (no external light source was present during the experiments) and a homogeneous static background (all human-object interactions were performed on top of a desk covered with a green tablecloth). Snapshots of the captured video streams from each viewpoint are depicted in Fig. \ref{fig:dataset}.

\begin{table}[t]
\begin{center}
\renewcommand*{\arraystretch}{1.05}
\resizebox{\linewidth}{!}{\begin{tabular}{l|c|c|c|c|c}
Dataset & Types & Affordances & Interactions & Subjects & Available \\
\hline
\cite{kluth} & 8 & 1 & n/a & n/a & no \\
\cite{kragic1} & 4 & 1 & 4 & Robot arm & no\\
\cite{kragic2} & 6 & 3 & 7 & 4 & no\\
\cite{castellini} & 7 & 5 & 13 & 20 & no\\
Introduced & \textbf{14} & \textbf{13} & \textbf{54} & \textbf{105} & \textbf{yes}\\
\hline
\end{tabular}}
\end{center}
\caption{Characteristics of sensorimotor object recognition corpora reported in the literature, compared to the presented one. Number of object types, affordances, human-object interactions, and subjects, as well as data public availability, are reported.}
\label{tbl:datasets}
\end{table}

\begin{table*}[t]
\footnotesize
\begin{center}
\renewcommand*{\arraystretch}{1.05}
\resizebox{\linewidth}{!}{\begin{tabular}{|l||c c c c c c c c c c c c c|}
\hline
\multirow{2}{*}{\textbf{Object types}}& \multicolumn{13}{|c|}{\textbf{Affordances}} \\ 
\cline{2-14}
& Grasp & Lift & Push & Rotate & Open & Hammer & Cut & Pour & Squeeze & Unlock & Paint & Write & Type\\
\hline
Ball & $\surd$ & $\surd$ & $\surd$ &  &  &  &  &  &  &  &  &  & \\
Book & $\surd$ & $\surd$ & $\surd$ &  & $\surd$ & $\surd$ &  &  &  &  &  &  & \\
Bottle & $\surd$ & $\surd$ & $\surd$ &  &  &  &  & $\surd$ &  &  &  &  & \\
Box & $\surd$ & $\surd$ & $\surd$ & $\surd$ & $\surd$ &  &  &  &  &  &  &  & \\
Brush & $\surd$ & $\surd$ &  &  &  &  &  &  &  &  & $\surd$ &  & \\
Can & $\surd$ & $\surd$ & $\surd$ &  &  &  &  &  &  &  &  &  & \\
Cup & $\surd$ & $\surd$ & $\surd$ & $\surd$ &  &  &  &  &  &  &  &  & \\
Hammer & $\surd$ & $\surd$ &  &  &  & $\surd$ &  &  &  &  &  &  & \\
Key & $\surd$ & $\surd$ &  &  &  &  & $\surd$ &  &  & $\surd$ &  &  & \\
Knife & $\surd$ & $\surd$ &  &  &  &  & $\surd$ &  &  &  &  &  & \\
Pen & $\surd$ & $\surd$ &  &  &  &  &  &  &  &  &  & $\surd$ & \\
Pitcher & $\surd$ & $\surd$ & $\surd$ & $\surd$ &  &  &  & $\surd$ &  &  &  &  & \\
Smartphone & $\surd$ & $\surd$ & $\surd$ &  &  &  &  &  &  &  &  &  & $\surd$ \\
Sponge & $\surd$ & $\surd$ & $\surd$ & $\surd$ &  &  &  &  & $\surd$ &  &  &  & \\
\hline
\end{tabular}}
\end{center}
		\caption{Supported object and affordance types in the presented corpus. Considered object-affordance combinations are marked with $\surd$. }
\label{tbl:affordances}
\end{table*}

Regarding the nature of the supported human-object interactions, a set of 14 object types was considered (each type having two individual instantiations, \eg small and big ball). The appearance characteristics of the selected object types varied significantly, ranging from distinct shapes (like ``Box" or ``Ball") to more challenging ones (like ``Knife"). Taking into account the selected objects, a respective set of 13 affordance types was defined, covering typical manipulations of the defined objects. Concerning the complexity of the supported affordances, relatively simple (\eg ``Grasp"), complex (\eg leading to object deformations, like affordance ``Squeeze"), and continuous-nature ones (\eg affordance ``Write") were included. In contrast, other experimental settings in the literature have mostly considered simpler and less time evolving affordances, like ``Grasp" and ``Push". In Table \ref{tbl:affordances}, all supported types of objects and affordances, as well as all combinations that have been considered in the dataset, are provided. As listed, a total of $54$ object-affordance combinations (\ie human-object interactions) are supported. All participants were asked to execute all object-affordance combinations indicated in Table \ref{tbl:affordances} at least once. The experimental protocol resulted in a total of 20,830 instances, considering the data captured from each Kinect as a different human-object interaction instance. The length of every recording varied between 4 and 8 seconds.

\section{Sensorimotor Object Recognition} \label{sensorimotor}
We now proceed to describe the proposed sensorimotor object recognition system. Specifically, we first provide its overview, followed by details of video data processing and the DL modeling approaches considered.

\subsection{System overview}
The proposed system is depicted in Fig \ref{fig:visual}. Initially, the collected data are processed by the visual front-end module. This produces three visual feature streams, one of which corresponds to conventional object appearance, while the remaining two capture object affordance information. These streams are subsequently fed to appropriate DL architectures, implementing single-stream processing systems for the recognition of object types and affordances. Eventually, appearance and affordance information are combined to yield improved object recognition, following various fusion strategies.

\subsection{Visual front-end} \label{front}
The RGB data stream is initially mapped to the depth stream for each Kinect, making use of a typical calibration model \cite{herrera}. Since the exact positioning of the Kinect sensors is known in the developed capturing framework, a ``3D volume of interest" is defined for each Kinect, corresponding to the 3D space above the desk that includes all performed human-object interactions. Pixels that correspond to 3D points outside the defined volumes of interest are considered as background and the respective RGB/depth values are set to zero. Subsequently, a centered rectangular region ($300\times 300$ resolution), containing the observed object manipulations, is cropped from the aligned RGB and depth frames. Then, using a simple thresholding technique in the HSV color space \cite{vez}, pixels corresponding to the desk plane (tablecloth) are removed, and subsequently skin color pixels (corresponding to the hand of the performing subject) are separated from the object ones. 

\begin{figure*}[!t]
\begin{center}
	\includegraphics[width=\linewidth]{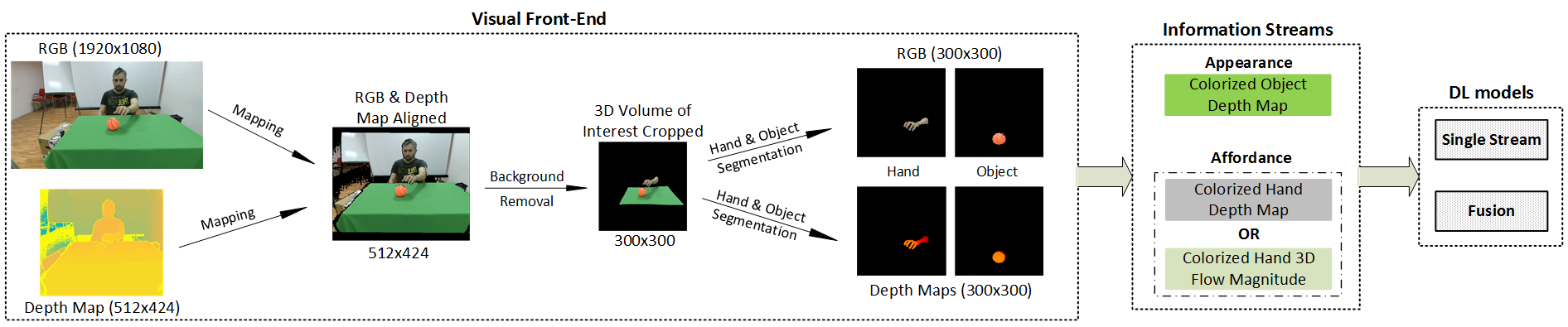}
\end{center} 
   \caption{System overview. The visual front-end module (left) processes the captured data, providing three information streams (middle) that are then fed into a single-stream or fusion DL model (right).}
\label{fig:visual}
\end{figure*}

\begin{figure}[!b]
\begin{center}
	\includegraphics[width=\linewidth]{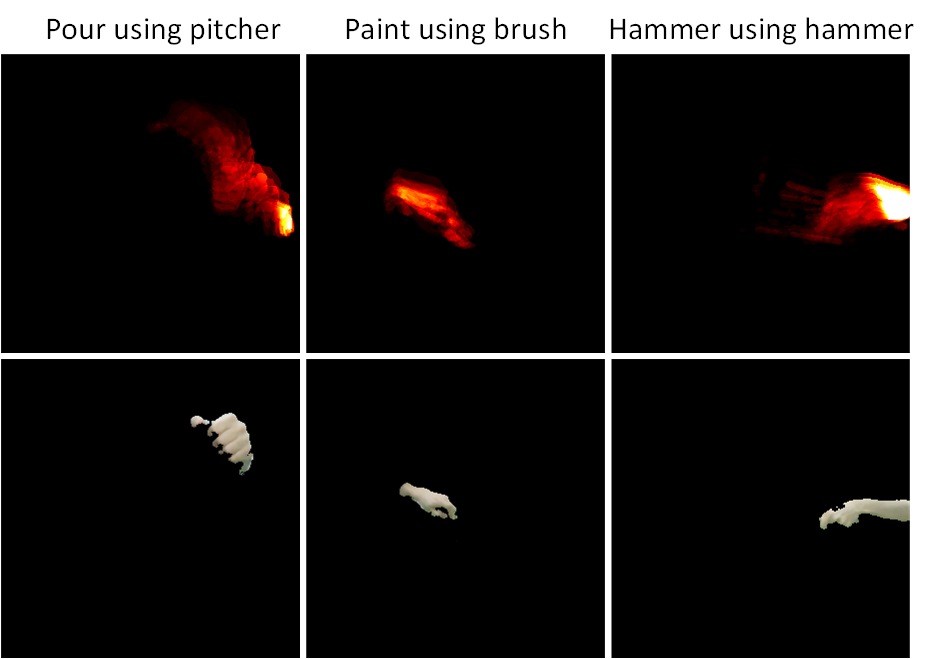}
\end{center} 
   \caption{ Examples of colorized flow magnitude field (top row) and RGB snapshots of the corresponding actions (bottom row).}
\label{fig:magn}
\end{figure}

For the extracted object and hand depth maps, a ``depth colorization" approach is followed, similar to the one introduced by Eitel \etal~\cite{eitel}. The depth colorization enables the common practice of utilizing networks (transfer learning \cite{razavian, yosinski}) pre-trained on ImageNet \cite{deng}, and fine-tuning them on the collected data. In particular, the depth value at every pixel location is linearly normalized in the interval $[0,255]$, taking into account the minimum and maximum depth values that have been measured in the whole dataset and for all pixel locations. Subsequently, a ``hot colormap" is applied for transforming each scalar normalized depth value to a triplet RGB one. In parallel, the 3D flow magnitude of the extracted hand depth maps is also computed. The algorithm of \cite{jaimez} for real-time dense RGB-D scene flow estimation is used. Denoting by $F_{t}(x,y,z)$ the 3D flow-field of the depth video at frame $t$, the 3D magnitude field $\sum_{t=1}^{T} \lvert F_{t}(x,y,z)\rvert$  is considered, accumulated over the video duration ($T$ frames). For the latter field, the same colorization approach with the RGB case is applied (Fig. \ref{fig:magn}). Thus, the visual front-end provides three information streams (Fig \ref{fig:visual} middle): a) colorized object depth maps, b) colorized hand depth maps, and c) colorized hand 3D flow magnitude fields. 

\subsection{Single-stream modeling} \label{single} 
\begin{figure}[!b]
\begin{center}
	\includegraphics[width=\linewidth]{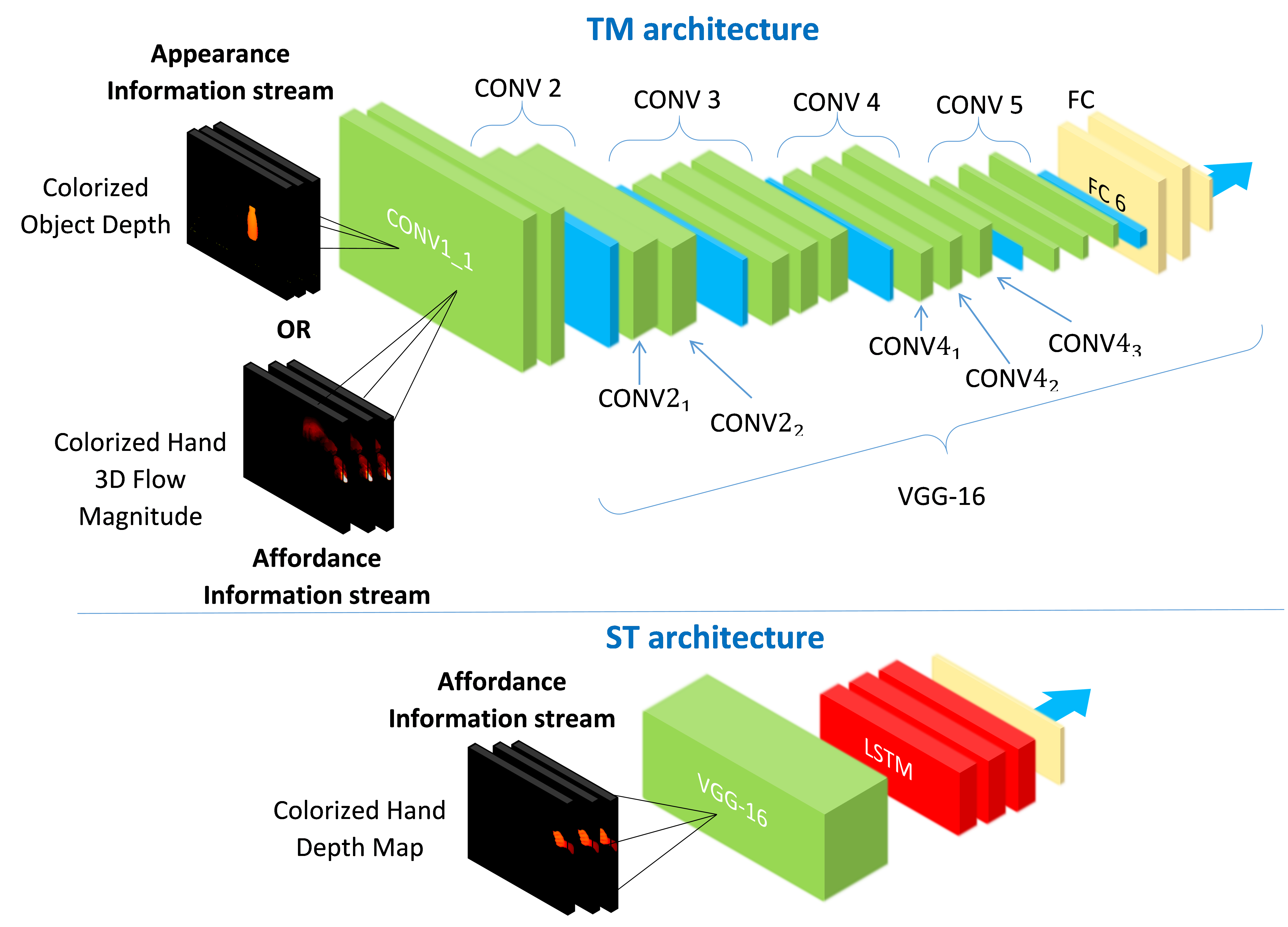}
\end{center} 
   \caption{Single-stream models. Top: appearance CNN for object recognition, and affordance CNN (TM architecture). Bottom: affordance CNN-LSTM (ST architecture). The CNN layer notation used in this paper is depicted at the top figure.}
\label{fig:single_stream}
\end{figure}

For each information stream, separate NN architectures are designed, as depicted in Fig \ref{fig:single_stream}. Regarding the \textbf{appearance stream}, the well-known VGG-16 network \cite{simonyan}, which consists of 16 layers in total, is used for analyzing the appearance of the observed objects. The VGG-16 model consists of 5 groups of Convolutional (CONV) layers and 3 Fully Connected (FC) ones. After each CONV or FC layer, a Rectified Linear Unit (RL) one follows. For the rest of the paper, the notation depicted in Fig \ref{fig:single_stream} (top) is used (\eg CONV$4_3$ is the $3^{rd}$ CONV layer of the $4^{th}$ group of convolutions). 

Regarding the \textbf{affordance stream}, the colorized hand depth map and 3D flow magnitude are alternatively used for encoding the corresponding dynamics. In particular, two distinct NN architectures, aiming at modeling different aspects of the exhibited motor (hand) actions, are designed: a) a ``Template-Matching" (TM), and b) a ``Spatio-Temporal" (ST) one. The development of the TM architecture is based solely on the use of CNNs (Fig \ref{fig:single_stream} top), aiming at estimating complex multi-level affordance-related patterns along the spatial dimensions. The different CONV layers of the employed CNN now model affordance-related patterns of increasing spatial complexity. With respect to the development of the ST architecture, a composite CNN (VGG-16) - Long-Short Term Memory (LSTM) \cite{lstm} NN is considered, where the output of a CNN applied at every frame is subsequently provided as input to an LSTM. This architecture is similar to the generalized ``LRCN" model presented in \cite{donahue2}. The aim of the ST architecture (Fig \ref{fig:single_stream} bottom) is to initially model correlations along the spatial dimensions and subsequently to take advantage of the LSTM sequential modeling efficiency for encoding the temporal dynamics of the observed actions. In preliminary experiments, the colorized hand depth map led to better results than using 3D flow information as input. A set of $20$ frames, uniformly sampled over the whole duration of the observed action, are provided as input to the LSTM.

\begin{figure*}[!t]
\begin{center}
	\includegraphics[width=.9\linewidth]{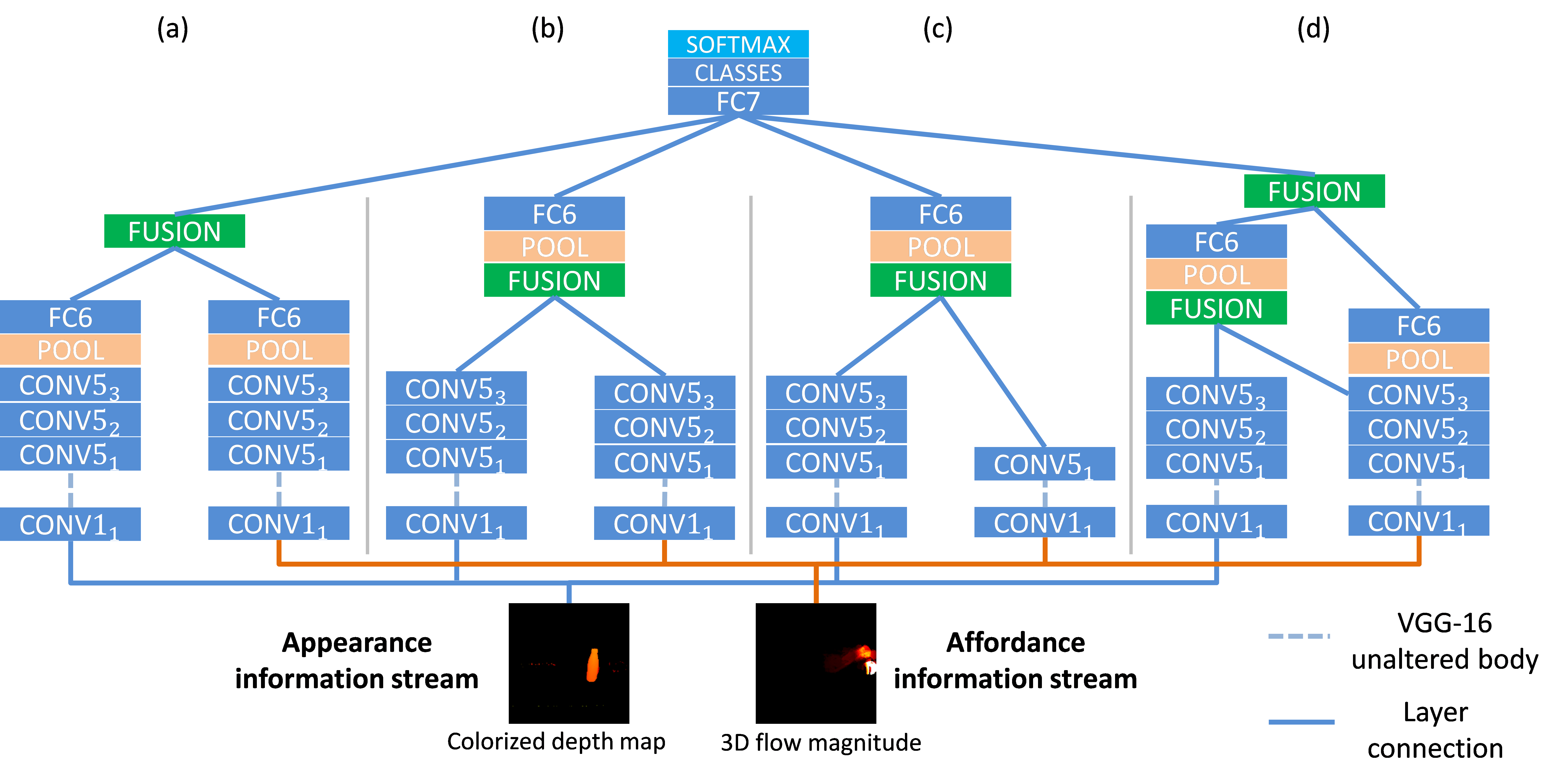}
\end{center} 
   \caption{Detailed topology of the GTM architecture for: a) late fusion at FC layer, b) late fusion at last CONV layer, c) slow fusion, and d) multi-level slow fusion. In each case, the left stream represents the appearance and the right the affordance network, respectively.}
\label{fig:gtm}
\end{figure*}
\subsection{Fusion architectures} \label{fusion}
Prior to the detailed description of the evaluated sensorimotor information fusion principles, it needs to be noted that these are implemented within two general NN architectures, namely the ``Generalized Template-Matching" (GTM) and the ``Generalized Spatio-Temporal" (GST) one. The GTM (Fig. \ref{fig:gtm}) and the GST (Fig. \ref{fig:gst}) architectures are derived from the corresponding TM and ST ones, respectively, and their fundamental difference concerns the nature of the affordance stream modeling; GTM focuses on modeling correlations along the spatial dimensions, while GST relies on encoding time-evolving procedures of the performed human actions. Anatomical studies \cite{brandi} on the physiological interconnections between the ventral and the dorsal streams have resulted, among others, in the following dominating hypothesis: The ventral (appearance) stream might receive up-to-date action-related information from the dorsal (affordance) stream, in order to refine the object internal representation \cite{neuro1}.

\subsubsection{Late fusion}
Late fusion refers to the combination of information at the end of the processing pipeline of each stream. For the GTM architecture, this is implemented as the combination of the features at: a) the same FC layer, or b) the last CONV layer. The FC layer fusion is performed by concatenating the FC features of both streams. It was experimentally shown that fusion after the RL6 layer was advantageous, compared to concatenating at the output of the FC6 layer (\ie before the nonlinearity). After fusion, a single processing stream is formed (Fig. \ref{fig:gtm}a). Regarding fusion at the last CONV layer, the RL$5_3$ activations of both appearance and affordance CNNs are stacked. After feature stacking, a single processing stream is again formed with four individual structural alternatives, using: i) 1 CONV ($1 \times1$ kernel size) and 1 FC, ii) 2 CONV ($1 \times1$ kernel size) and 1 FC, iii) 1 CONV ($1 \times1$ kernel size) and 2 FC (best performance, depicted in Fig. \ref{fig:gtm}b), and iv) 2 CONV ($1 \times1$ kernel size) and 2 FC layers. 

For the GST architecture, the late fusion scheme considers only the concatenation of the features of the last FC layers of the appearance CNN and the affordance LSTM model, as depicted in Fig. \ref{fig:gst}a. In particular, the features of the FC7 layer of the appearance CNN and the internal state vector [\textbf{$h(t)$}] of the last LSTM layer of the affordance stream are concatenated at every time instant (\ie at every video frame). Eventually, a single stream with 2 FC layers is formed. On the other hand, there is accumulated evidence that asynchronous communication and feedback loops occur during the sensorimotor object recognition process \cite{neuro2}. In this context, an asynchronous late fusion approach is also investigated for the GST architecture. Specifically, the GST late fusion scheme (Fig. \ref{fig:gst}a) is again applied. However, the information coming from the affordance stream [\ie the internal state vector $h(t)$ of the last LSTM layer] is provided with a time-delay factor, denoted by $\tau >0$, compared to the FC features of the appearance stream; in other words, the features of the affordance stream at time $t-\tau$ are combined with the appearance features at time $t$.

\subsubsection{Slow fusion}
\begin{figure}[t]
\begin{center}
	\includegraphics[width=\linewidth]{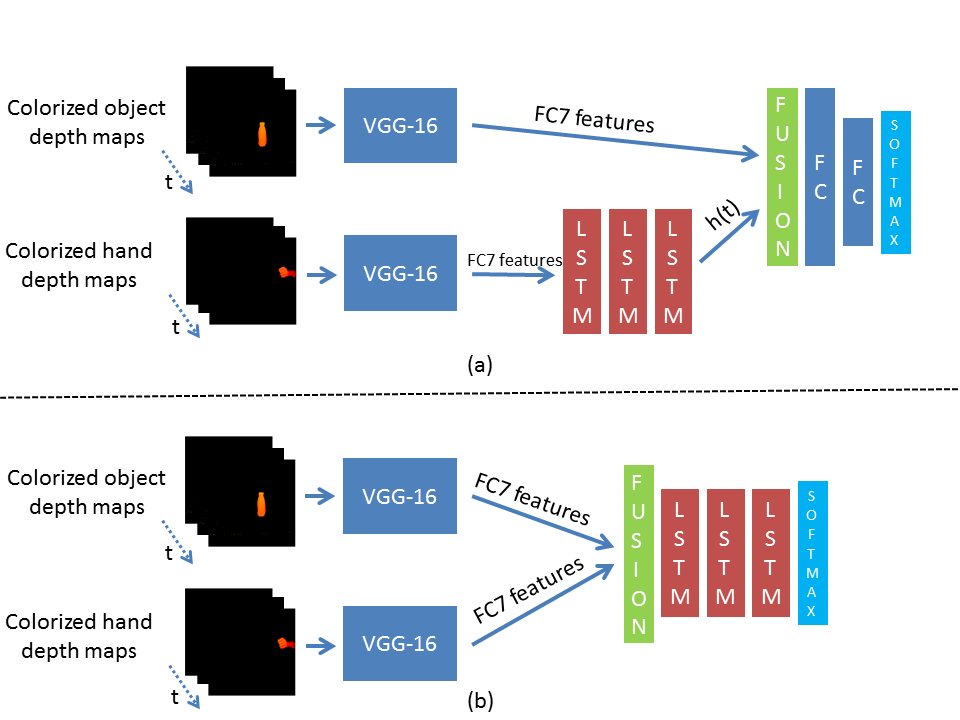}
\end{center} 
   \caption{Detailed topology of the GST architecture for: a) late fusion and b) slow fusion.}
\label{fig:gst}
\end{figure}

Slow fusion for the GTM architecture corresponds to the case of combining the CONV feature maps of the appearance and affordance CNNs in an intermediate layer (\ie not the last CONV layer) and subsequently forming a single processing stream, as depicted in Fig. \ref{fig:gtm}c. For realizing this, two scenarios are considered, which correspond to the fusion of information from the two aforementioned CNNs at different levels of granularity: a) combining the feature maps of the appearance and the affordance CNN from the same layer level; and b) combining the feature maps of the appearance and the affordance CNN from different layer levels. The actual fusion operator is materialized by simple stacking of the two feature maps. It needs to be noted that only appearance and affordance feature maps of same dimensions are combined. For the GST architecture, the slow fusion scheme considers only the concatenation of the features of the RL7 layer of the appearance and the affordance CNNs models, followed by an LSTM model, as can be seen in Fig. \ref{fig:gst}b.

In order to simulate the complex information exchange routes at different levels of granularity between the two streams, a multi-level slow fusion scheme is also examined. In particular, the two streams are connected both at an intermediate/last CONV and at the FC layers. The particular NN topology that implements this multi-level slow fusion scheme for the GTM architecture is illustrated in Fig. \ref{fig:gtm}d.

In the remainder of the paper, the following naming convention is used for describing the different proposed NN architectures: GAT$_{FT}(param)$, where the Generalized Architecture Type, GAT $\in$ \{GTM, GST\}, and the Fusion Type, FT $\in \{LS, LA, SSL, SML\}$ $\equiv$ \{Late Synchronous, Late Asynchronous, Slow Single Level, Slow Multi Level\} and $param$ indicates the specific parameters for each particular fusion scheme (as detailed above). At this point, it needs to be highlighted that any further information processing performed in the affordance stream after the fusion step does not contribute to the object recognition process; hence, it is omitted from the descriptions in this work. 

\section{Experimental Results} \label{results}

The proposed NN architectures are evaluated using the introduced dataset. The involved human subjects were randomly divided into training, validation, and test sets (25\%, 25\%, and 50\%). The utilized VGG-16 network was pre-trained on ImageNet. For all $300\times 300$ formed frames, a $224\times 224$ patch was randomly cropped and provided as input to the NNs. The negative log-likelihood criterion was selected during training, whereas for back-propagation, Stochastic Gradient Descent (SGD) with momentum set equal to $0.9$ was used. The GTM- and GST-based NNs were trained with learning rate set to $5\times 10^{-3}$ (decreasing by a factor of $5\times 10^{-1}$ when the validation error curve plateaued) for 60 and 90 epochs, respectively. For the implementation, the \textit{Torch}\footnote{http://torch.ch/} framework and a Nvidia Tesla K-40 GPU were used.

\subsection{Single-stream architecture evaluation}
The first set of experiments concerns the evaluation of the single-stream models (Section \ref{single}). From the results presented in Table \ref{tbl:single} (only overall classification accuracy is given), it can be observed that object recognition using only appearance information yields satisfactory results (85.12\% accuracy). Regarding affordance recognition, the TM architecture outperforms the ST one, indicating that the CNN model encodes the motion characteristics more efficiently than the composite CNN-LSTM one. For the ST model 3 LSTM layers with 4096 hidden units each were used, based on experimentation.

\subsection{GTM and GST architectures evaluation}

In Table \ref{tbl:gtm}, evaluation results from the application of different GTM-based fusion schemes (Section \ref{fusion}) are given. From the presented results, it can be seen that for the case of late fusion combination of CONV features (\ie fusion at the RL$5_{3}$ layer) is generally advantageous, since the spatial correspondence between the appearance and the affordance stream is maintained. Concerning single-level slow fusion models, different models are evaluated. However, single-level slow fusion tends to exhibit lower recognition performance than late fusion. Building on the evaluation outcomes of the single-level slow and late fusion schemes, multi-level slow fusion architectures are also evaluated. Interestingly, the GTM$_{SML}$(RL5$_{3}^{app}$, RL5$_{3}^{aff}$, RL6) outperforms all other GTM-based models. This is mainly due to the preservation of the spatial correspondence (initial fusion at the CONV level), coupled with the additional correlations learned by the fusion at the FC level. 

Experimental results from the application of the GST-based fusion schemes (Section \ref{fusion}) are reported in Table \ref{tbl:gst}. In all cases, a set of $20$ uniformly selected frames were provided as input to the respective NN. Additionally, two evaluation scenarios were realized, namely when for the final object classification decision the prediction of only the last frame was considered (``last-frame") or when the predictions from all frames were averaged (``all-frames"). For the case of the synchronous late fusion, it can be seen that the averaging of the predictions from all frames is advantageous. Concerning the proposed asynchronous late fusion scheme, evaluation results for different values of the delay parameter $τ$ are given. It can be observed that asynchronous fusion leads to decreased performance, compared to the synchronous case, while increasing values of the delay parameter $\tau$ lead to a drop in the recognition rate. Moreover, the slow fusion approach results to a significant decrease of the object recognition performance.

\begin{figure}[t]
\begin{center}
	\includegraphics[width=\linewidth]{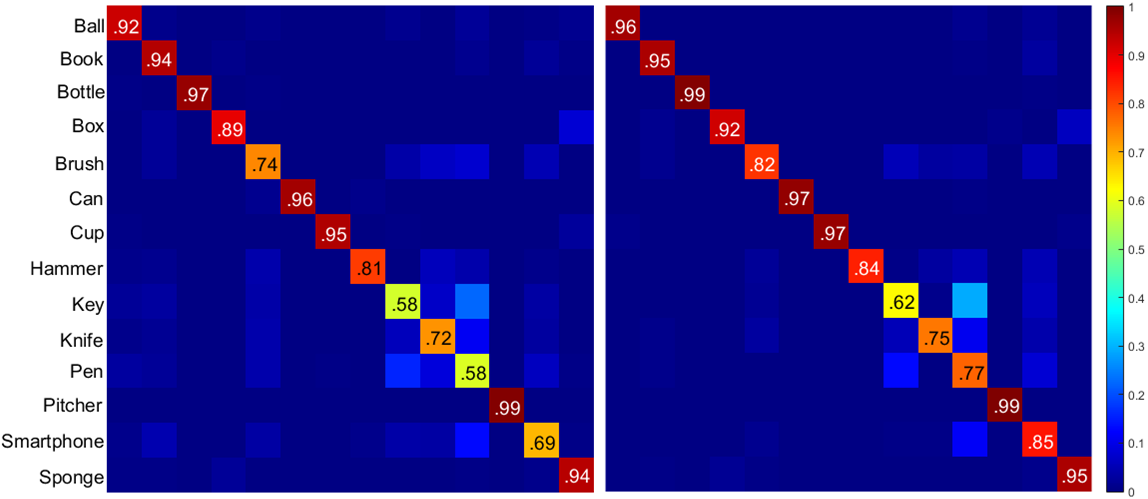}
\end{center} 
   \caption{Object recognition confusion matrices of appearance CNN (left) and GTM$_{SML}$(RL5$_{3}^{app}$, RL5$_{3}^{aff}$, RL6) architecture (right).}
\label{fig:heat}
\end{figure}

From the presented results, it can be observed that the GTM$_{SML}$(RL5$_{3}^{app}$, RL5$_{3}^{aff}$, RL6) architecture constitutes the best performing scheme. The latter achieves an absolute increase of $4.31\%$ in the overall recognition performance (which corresponds to an approximately $~29\%$ relative error reduction), compared to the appearance CNN model (baseline method). For providing a better insight, the object recognition confusion matrices obtained from the application of the GTM$_{SML}$(RL5$_{3}^{app}$, RL5$_{3}^{aff}$, RL6) architecture and the appearance CNN are given in Fig. \ref{fig:heat}. From the presented results, it can be observed that the proposed fusion scheme boosts the performance of all supported object types. This demonstrates the discriminative power of affordance information. Additionally, it can be seen that objects whose shape cannot be efficiently captured (\eg small-size ones like ``Pen", ``Knife", ``Key", etc.) are favored by the proposed approach. Moreover, affordance information is also beneficial for objects that exhibit similar appearance (\eg ``Brush" with ``Pen" and ``Knife").

\begin{table}[t]
\begin{center}
\resizebox{.95\linewidth}{!}{\begin{tabular}{l|c|c}
Method & Task & Accuracy (\%) \\
\hline
Appearance CNN & object recognition & 85.12 \\
Affordance CNN & affordance recognition & 81.92 \\
Affordance CNN-LSTM & affordance recognition & 69.27 \\
\hline
\end{tabular}}
\end{center}
\caption{Single-stream results for object and affordance recognition.}
\label{tbl:single}
\end{table}

\begin{table}[t]
\begin{center}
\resizebox{.95\linewidth}{!}{\begin{tabular}{l|c}
GTM-based fusion architecture [after fusion] & Accuracy (\%) \\
\hline
\\[-1em]
GTM$_{LS}$(FC6) & 87.40 \\
GTM$_{LS}$(RL5$_{3}$) [1 CONV, 1 FC] & 87.65\\
GTM$_{LS}$(RL5$_{3}$) [1 CONV, 2 FC] & 88.24\\
GTM$_{LS}$(RL5$_{3}$) [2 CONV, 1 FC] & 87.64\\
GTM$_{LS}$(RL5$_{3}$) [2 CONV, 2 FC] & 86.40\\
\hline
\\[-1em]
GTM$_{SSL}$(RL3$_{3}^{app}$, RL3$_{3}^{aff}$) & 78.74\\
GTM$_{SSL}$(RL4$_{3}^{app}$, RL4$_{3}^{aff}$) & 87.20\\
GTM$_{SSL}$(RL4$_{3}^{app}$, RL4$_{1}^{aff}$) & 85.82\\
GTM$_{SSL}$(RL5$_{3}^{app}$, RL5$_{1}^{aff}$) & 88.13\\
\hline
\\[-1em]
GTM$_{SML}$(RL5$_{3}^{app}$, RL5$_{1}^{aff}$, RL6) & 88.23\\
GTM$_{SML}$(RL5$_{3}^{app}$, RL5$_{3}^{aff}$, RL6) & \textbf{89.43}\\
\hline
\end{tabular}}
\end{center}
\caption{Object recognition results using different GTM-based fusion schemes.}
\label{tbl:gtm}
\end{table}

\subsection{Comparison with probabilistic fusion}
The GTM$_{SML}$(RL5$_{3}^{app}$, RL5$_{3}^{aff}$, RL6) architecture is also comparatively evaluated, apart from the appearance CNN model, with the following typical probabilistic fusion approaches of the literature: a) the product rule for fusing the appearance and the affordance CNN output probabilities, b) concatenation of appearance and affordance CNN features and usage of a SVM classifier (RBF kernel) \cite{castellini,kragic2}, and c) concatenation of appearance and affordance CNN features and usage of a naive Bayes classifier \cite{kragic1}. From the results in Table \ref{tbl:prob}, it can be seen that the literature approaches for fusing the appearance and affordance information streams fail to introduce an increase in the object recognition performance in the introduced challenging dataset; the aforementioned methods were evaluated under significantly simpler experimental settings. On the contrary, the proposed approach exhibits a significant performance increase over the baseline method (appearance CNN). 

\begin{table}[t]
\begin{center}
\resizebox{.95\linewidth}{!}{\begin{tabular}{l|c}
GST-based fusion architecture [after fusion] & Accuracy (\%) \\
\hline
\\[-1em]
GST$_{LS}$(last-frame) & 86.28\\
GST$_{LS}$(all-frames) [1 CONV, 2 FC] & \textbf{86.50}\\
\hline
\\[-1em]
GST$_{LA}$(all-frames, $\tau =2$) & 86.42\\
GST$_{LA}$(all-frames, $\tau =4$) [1 CONV, 2 FC] & 86.17\\
GST$_{LA}$(all-frames, $\tau =6$) [1 CONV, 2 FC] & 85.28\\
\hline
\\[-1em]
GST$_{SSL}$(all-frames) [1 CONV, 2 FC] & 79.65\\
\hline
\end{tabular}}
\end{center}
\caption{Object recognition results using different GST-based fusion schemes.}
\label{tbl:gst}
\end{table}

\begin{table}[t]
\begin{center}
\resizebox{.9\linewidth}{!}{\begin{tabular}{l|c|c}
Fusion architecture & Fusion Layer & Accuracy (\%)\\
\hline
Appearance CNN & no fusion & 85.12\\ \hdashline 
Product Rule & Softmax & 73.45\\
SVM \cite{kragic2,castellini} & RL7 & 83.43\\
Bayes \cite{kragic1}& RL7 & 75.86\\
GTM$_{SML}$ & RL5$_{3}^{app}$, RL5$_{3}^{aff}$, RL6 & \textbf{89.43}\\
\hline
\end{tabular}}
\end{center}
\caption[caption]{Comparative evaluation of GTM$_{SML}$(RL5$_{3}^{app}$, RL5$_{3}^{aff}$, RL6) architecture.}
\label{tbl:prob}
\end{table}

\section{Conclusions} \label{conclusion}

In this paper, the problem of sensorimotor 3D object recognition following the deep learning paradigm was investigated. A large public 3D object recognition dataset was also introduced, including multiple object types and a significant number of complex affordances, for boosting the research activities in the field. Two generalized neuro-biologically and neuro-physiologically grounded neural network architectures, implementing multiple fusion schemes for sensorimotor object recognition were presented and evaluated. The proposed sensorimotor multi-level slow fusion approach was experimentally shown to outperform similar probabilistic fusion methods of the literature. Future work will investigate the use of NN auto-encoders for modeling the human-object interactions in more details and the application of the proposed methodology to more realistic, ``in-the-wild" object recognition data.

{\small
\bibliographystyle{ieee}
\bibliography{egbib}
}

\end{document}